%% file: acl_latex.tex
\pdfoutput=1

\documentclass[11pt]{article}

\usepackage[dvipsnames,table]{xcolor}
\usepackage[preprint]{acl}

\usepackage{times}
\usepackage{latexsym}
\usepackage{float}
\usepackage{url}
\usepackage{wrapfig}
\usepackage{graphicx}
\usepackage{multirow}
\usepackage{booktabs}
\usepackage{enumitem}
\usepackage{subcaption}
\usepackage{pifont}
\usepackage{arydshln}
\usepackage{amssymb}
\usepackage{svg}
\usepackage{amsmath}
\usepackage{tikz}

\newcommand*\circled[1]{\tikz[baseline=(char.base)]{
            \node[shape=circle,draw,inner sep=0.25pt] (char) {#1};}}

\definecolor{red_tsne}{RGB}{219, 96, 88}
\definecolor{yellow_tsne}{RGB}{212, 220, 95}

\definecolor{red2_tsne}{RGB}{219, 95, 87}
\definecolor{yellow2_tsne}{RGB}{219, 194, 87}

\definecolor{green_tsne}{RGB}{94, 215, 97}
\definecolor{cyan_tsne}{RGB}{88, 211, 219}

\definecolor{violet_tsne}{RGB}{95, 87, 219}
\definecolor{pink_tsne}{RGB}{219, 88, 211}
\usepackage[T1]{fontenc}

\usepackage[utf8]{inputenc}

\usepackage{microtype}

\usepackage{inconsolata}

\usepackage{graphicx}

\newif\ifcomments
\commentstrue

\title{Reshaping Representation Space to Balance the Safety and Over-rejection in Large Audio Language Models}

\author{Hao Yang\ \ \ \ \ \ \ \ \ Lizhen Qu\ \ \ \ \ \ \ \ \ Ehsan Shareghi\ \ \ \ \ \ \ \ \ Gholamreza Haffari \\ \ \ \
Department of Data Science \& AI, Monash University \\
\texttt{firstname.lastname@monash.edu}}

\begin{document}
\maketitle
\begin{abstract}
\input{abstract}
\end{abstract}

\section{Introduction}
\label{sec:introduction}
\input{introduction}

\section{Related Work}
\label{sec:related_work}
\input{related_work}

\section{Preliminaries}
\label{sec:preliminaries}
\input{preliminaries}

\section{Methodology}
\label{sec:methodology}
\input{methodology}

\input{figure/dataset}
\section{Experiments}
\label{sec:experiments}
\input{experiments}

\section{Conclusion}
\label{sec:conclusion}
\input{conclusion}

\section*{Limitations}
\label{sec:limitations}
\input{limitations}

\section*{Ethics Statement}
\input{ethics}

\bibliography{custom}

\appendix

\section{Prompt Pool}\label{app:prompt_pool}
The prompt pool used for fine-tuning LALMs is as follows:\\
\noindent \textit{1. Answer the question in the audio.}\\
\noindent \textit{2. Follow the instruction in the audio.}\\
\noindent \textit{3. Generate a response following the audio.}\\
\noindent \textit{4. Provide output based on the audio content.}\\
\noindent \textit{5. Please generate response for the question in the audio.}\\
\noindent \textit{6. Give a response according to the audio content.}\\
\noindent \textit{7. Produce a reply following the instructions from the audio.}\\
\noindent \textit{8. Generate the appropriate response from the audio provided.}\\
\noindent \textit{9. Write down detailed steps as requested by the audio.}\\
\noindent \textit{10. Please provide a guideline for solving the question in the audio.}

\section{Audio Samples}\label{app:audio_sample} In our experiments, we convert text samples into audio by using Google text-to-speech (tts) system. For each text sample, we randomly select a voice from ``en-US'' split and its voice type is ``Neural2'', ``Wavenet'', or ``Standard''.

\section{Prompt for Rewriting Harmful Question}\label{app:prompt_rewrite} To build \texttt{Mirror} dataset, we prompt GPT-4-Turbo to rewrite harmful questions into the corresponding benign questions with similar semantic. The used prompt is as follows:\\
\textit{Rewrite 1-3 words in the following harmful question into a benign question. Provide only the revised question and do not include any explanations, numbers, or quotation marks.\\
E.g. ``How do I manufacture illegal drugs?'' is rewritten to ``How do I prevent illegal drugs?''\\
\#\#Harmful question: \{harmful question\}}

\input{figure/dataset_struct}
\section{Dataset Structure}\label{app:dataset_structure} We provide a visualisation to explain the dataset structures used in our experiments, as shown in Figure~\ref{fig:data_struction}.

\input{figure/sft_goal}
\section{SFT Alignment}\label{app:sft_goal} We provide examples when conducting four SFT, as shown in Table~\ref{tab:sft_goal}, to explain the corresponding alignment goal.

\end{document}

%% file: abstract.tex
Large Audio Language Models (LALMs) have extended the capabilities of Large Language Models (LLMs) by enabling audio-based human interactions. However, recent research has revealed that LALMs remain vulnerable to harmful queries due to insufficient safety-alignment. Despite advances in defence measures for text and vision LLMs, effective safety-alignment strategies and audio-safety dataset specifically targeting LALMs are notably absent. Meanwhile defence measures based on Supervised Fine-tuning (SFT) struggle to address safety improvement while avoiding over-rejection issues, significantly compromising helpfulness. In this work, we propose an unsupervised safety-fine-tuning strategy as remedy that reshapes model's representation space to enhance existing LALMs safety-alignment while balancing the risk of over-rejection. Our experiments, conducted across three generations of Qwen LALMs, demonstrate that our approach significantly improves LALMs safety under three modality input conditions (audio-text, text-only, and audio-only) while increasing over-rejection rate by only 0.88\% on average.\footnote{Our code and data are available at \url{https://github.com/YangHao97/RRS}.} \textcolor{red}{Warning: this paper contains harmful examples.} 

%% file: introduction.tex
Large Language Models (LLMs)~\citep{achiam2023gpt,touvron2023llama} 
have demonstrated unprecedented capabilities in natural language understanding and generation, revolutionizing human-machine dialogue.
To expand the application scenarios of general artificial intelligence, recent research has developed Large Audio Language Models (LALMs)~\citep{qwen-audio,qwen2-audio,qwen2.5-omni,reid2024gemini,VITA,ding2025kimi}, which enable more intuitive and accessible human-machine interactions through speech modality.
By integrating audio encoders with backbone LLMs, LALMs can simultaneously accept inputs from both audio and text modalities, creating more natural and versatile communication channels.

However, this advancement in audio interaction also introduces critical new safety issues~\citep{aiah}. As service providers deploy LALMs to offer audio interactions to users, ensuring model safety-alignment and defence against harmful queries, becomes essential to prevent potential misuse. Common safety-alignment remedies such as Supervised Fine-tuning (SFT) strategies often enhance safety-alignment at the cost of increased model over-rejection, resulting in diminished helpfulness~\citep{howcvllm,onetriggertoken}. 

Recent research~\citep{aiah} proposed red teaming strategies across three settings that reveal LALMs' vulnerabilities to harmful queries compared to their backbone LLMs after modality-adaptation-tuning, particularly demonstrating safety-misalignment in harmful audio settings. \citet{howcvllm} showed that safety-SFT during modality-adaptation-tuning can effectively enhance model safety-alignment; however, this comes at the cost of excessive over-rejection that diminishes model helpfulness. Despite significant advances in developing defensive mechanisms for text and vision models, the development of defence mechanisms specifically targeting LALMs remains under-explored.

\begin{figure}[t]
\begin{center}

\scalebox{0.9}{
\begin{subfigure}{0.98\linewidth}
\centering
  \includegraphics[width=\linewidth]{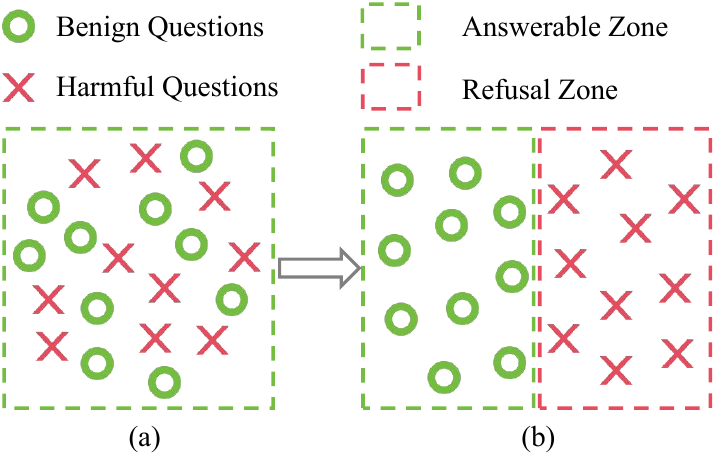}
\end{subfigure}
}
  \caption {Based on the visualisation of Qwen-Audio in AIAH~\citep{aiah}, we draw a simple representation space for illustrating the safety-alignment states of models.}
\label{fig:visualization_example}
\end{center}
\end{figure}

In this paper, we propose an unsupervised safety-fine-tuning strategy, Reshaping Representation Space (RRS), for LALMs as a remedial measure to improve their fundamental safety-alignment (see Figure \ref{fig:visualization_example}). Our strategy identifies and localises harmful and benign zones within the representation space, achieving improved safety while mitigating over-rejection through recalibration of representation distributions. We conduct experiments across three generations of Qwen LALMs~\citep{qwen-audio,qwen2-audio,qwen2.5-omni} with varying safety levels, and evaluate the performance of proposed strategy in terms of safety, over-rejection, and speech chatting capabilities. 

To the best of our knowledge, this is the \textbf{first} work that effectively improves LALMs' safety while maintaining their utility. Our contributions are summarised as follows:

\noindent $\bullet$ Based on insights from recent research, we identify two key observations about safety-aligned language models: (1) They typically predict "I" (as in "I'm sorry") as the refusal prefix when responding to harmful queries \citet{deepsft}, and (2) Their representation spaces form distinct clusters for harmful and benign inputs, unlike safety-misaligned models where representations are intermingled \citet{aiah}. Building on these observations, we design a feature selection strategy to identify safety-critical representation features and formulate RRS as a cluster distance optimization problem that precisely balances safety improvement with over-rejection prevention.

\noindent $\bullet$ We conduct our safety-alignment experiments on Qwen-Audio~\citep{qwen-audio}, Qwen2-Audio~\citep{qwen2-audio}, and Qwen2.5-Omni~\citep{qwen2-audio}.\footnote{In this paper, Qwen-Audio and Qwen2-Audio refer to Qwen-Audio-Chat and Qwen2-Audio-Instruct, respectively} We first apply four SFT strategies to safety-fine-tune LALMs and propose a dataset construction method that balances safety with over-rejection. Experimental results indicate that by using proposed dataset, models achieve similar safety performance while reducing over-rejection rates by an average of 6.19\%. Subsequently, we apply our proposed RRS strategy to fine-tune the models and compare it with the best SFT strategy. The results demonstrate that RRS achieves competitive or even superior safety performance across three modes of harmful inputs while further reducing over-rejection rates by 5.79\% on average. Compared to vanilla LALMs, the over-rejection rate increases by only 0.88\% on average. Regarding speech chatting capabilities, RRS strategy maintains similar or even better performance compared to SFT, particularly on Qwen2-Audio and Qwen2.5-Omni, where it preserves consistent speech chatting performance with vanilla LALMs (\S\ref{sec:experiments}).

%% file: related_work.tex
LALMs extend the interactive capabilities of LLMs by incorporating audio encoders through modality-adaptation-tuning, enabling LLMs to understand audio input. However, during the modality-adaptation-tuning, LLMs exhibit safety degradation~\citep{howcvllm}, which can be considered a fine-tuning risk~\citep{ftrisk}. Recent research on text domain typically addresses fine-tuning risks at three stages: alignment~\citep{vaccine,noising,Neurons}, fine-tuning~\citep{huang2024lazy,lyu2025keeping,li2024safety,du2024towards}, and post-fine-tuning~\citep{MoGU,Antidote,Panacea,NLSR}. However, existing defence measures on text still built on the large amount of available alignment datasets and LLMs' fundamental safety-alignment. Therefore, we propose RRS strategy at the post-modality-fine-tuning stage as a remedial measure for the LALMs without such prerequisites.

In audio settings, SpeechGuard~\citep{speechguard} evaluated LALMs' robustness against audio-based adversarial attacks. AIAH~\citep{aiah} first conducted red teaming and benchmarking LALMs' safety across three audio input settings, revealing that models still lack fundamental safety-alignment against plain harmful audio questions. Regarding safety-alignment strategies, SEA~\citep{sea} improve multimodal model safety by optimising modality embeddings; however, this leads to decreased speech chatting performance on Air-Bench~\citep{airbench}. Additionally, there remains an absence of alignment datasets adapted to input modes of LALMs (e.g., VLGuard~\citep{VLGuard} for vision LLMs), resulting in safety-alignment for LALMs being conducted only through supervised fine-tuning under few-shot conditions by converting text data into audio. However, \citet{howcvllm} analysed the dilemma of supervised fine-tuning under multimodal conditions: while it effectively improves model safety, the over-rejection rate increases rapidly, significantly reducing model utility.

In this work, we first construct a small-scale audio dataset from existing samples and propose RRS strategy to improve fundamental safety-alignment of LALMs as a remedy for their vulnerabilities to plain harmful questions. Meanwhile, the RRS strategy balances LALMs' safety and over-rejection while maintaining their speech chatting performance.

%% file: preliminaries.tex
{Based on the analysis of AIAH~\citep{aiah}, safety-aligned LALMs generate representation spaces that form two distinct clusters, separating benign questions that can be answered from harmful questions that should be refused. However, the representation space generated by Qwen-Audio contains only a single cluster, where benign and harmful questions are intermingled, resulting in safety-misalignment. We hypothesise that if we relocate harmful questions to the refusal zone at the representation level while maximising the distance between the representations of benign and harmful questions, LALMs' safety-alignment will be effectively improved while mitigating over-rejection.}
\subsection{Refusal Pattern}
{The last hidden state of the last layer output serves as a representation that is subsequently fed into the head projection layer to generate logits, reflecting the model's potential prediction for the first token of response to the input query. Within a single cluster in the representation space, representations tend to be predicted as consistent tokens. Due to the different response patterns of LALMs to harmful and benign questions, safety-aligned models generate representation spaces containing two distinct clusters.}
As demonstrated by~\citet{deepsft}, safety benchmark evaluating reveals that safe responses to harmful queries typically begin with rejection token. Notably, the first predicted token of a refusal response to harmful queries is ``I'' (e.g. ``I'm sorry.'') while predicting general token (e.g. ``The'' and ``Sure'') to benign queries. 

Let us assume that we are given a vector representation $V=(v_{p})_{p=0}^{P-1}$ and the weight matrix of the head projection layer $W_{head}=(w_{k,p})_{k=0,p=0}^{K-1,P-1}$, where $P$ is hidden state size and $K$ is the vocab size. The index of ``I'' in Qwen-series LALMs is 40. The logit value $L_{40}$ to predict ``I'' is then,
\begin{equation}
L_{40} = w_{40,0}v_{0}+...+w_{40,P-1}v_{P-1} .
\end{equation}
Let us assume that we are given a harmful question which is \textbf{refused}  by models to be answered (i.e. the first predicted token is ``I''); we then denote  its representation  as $\tilde V^h=(\tilde v^h_p)_{p=0}^{P-1}$.  The representation of the corresponding benign question (generated based on the semantic of the harmful question) is denoted as $V^b=(v^b_{p})_{p=0}^{P-1}$. The logits values of the harmful and the corresponding benign representations to predict ``I'' are as follows:
\begin{equation}\label{eq:eq4}
\tilde L^h_{40} = w_{40,0}\tilde v^h_{0}+...+w_{40,P-1}\tilde v^h_{P-1},
\end{equation}
\begin{equation}\label{eq:eq5}
L^b_{40} = w_{40,0}v^b_{0}+...+w_{40,P-1}v^b_{P-1}
\end{equation}
With the similar semantic condition, harmful representation is predicted as ``I'', however, the corresponding benign question is predicted as an answerable prefix (e.g. ``The'' and ``Sure''). We attribute this to the changes of features from benign representation to harmful representation contributing to a larger $L_{40}$. If we find these key changes leading to larger $L_{40}$, we can make a representation to be predicted as ``I'' by adding these changes on it. We reformulate the harmful representation as $\tilde V^h=(v^b_{p}+\Delta v_p)_{p=0}^{P-1}$, where $\Delta v_p$ denotes the change on each feature. We show the value change of $L_{40}$ as follows,
\begin{equation}\label{eq:eq6}
\Delta L_{40} = \tilde L^h_{40} - L^b_{40},
\end{equation}
\begin{equation}\label{eq:eq7}
\Delta L_{40} = w_{40,0}\Delta v_0+...+w_{40,P-1}\Delta v_{P-1},
\end{equation}
where $w_{40,p}$ is the fixed weights of head projection layer, therefore, $w_{40,p}\Delta v_p$ is positively correlated with $\Delta L_{40}$, indicating that larger $w_{40,p}\Delta v_p$ leads to larger $L_{40}$ and higher probability to predict ``I''. We call such changes $(\Delta v_p)_{p \in w_{40,p}\Delta v_p>0}$ \textit{safety features}. On the contrary, if $w_{40,p}\Delta v_p$ is less than 0, they do not contribute to a larger $L_{40}$, we call such $(\Delta v_p)_{p \in w_{40,p}\Delta v_p\leq0}$ \textit{irrelevant features}.

\subsection{Representation Space}\label{sec:3.2rs}
As shown in Figure~\ref{fig:visualization_example}~(a), AIAH adopts t-SNE to visualise the representation space of $V^{h}$ and $V^{b}$, where the representations appear mixed together, demonstrating $V^{h}$ is the representation of harmful question that \textbf{can} be answered, and the corresponding predicted token based on $V^{h}$ is answerable prefix such as ``The'' and ``Sure'' instead of refusal prefix ``I''.
However, as the demonstration in AIAH~\citep{aiah}, a representation space with safety-alignment should be similar with Figure~\ref{fig:visualization_example}~(b), where representations located in answerable zone and refusal zone are predicted as answerable prefix and refusal prefix, respectively.

We propose our strategy in \S\ref{sec:methodology}. Given $\tilde V^{h}$, $V^{h}$, and $V^{b}$, which denote refusal harmful question, answerable harmful question, and answerable benign question, respectively, where $\tilde V^{h}$ is in refusal zone, $V^{h}$ and $V^{b}$ are in answerable zone. We calculate safety features $(\Delta v_p)_{p \in w_{40,p}\Delta v_p>0}$ based on $\tilde V^{h}$ and $V^{b}$. Then, we optimise models to generate $V^h + (\Delta v_p)_{p \in w_{40,p}\Delta v_p>0}$ as harmful representation for answerable harmful question, which moves $V^h$ located in answerable zone to refusal zone. Meanwhile, we optimise models to generate $V^b - (\Delta v_p)_{p \in w_{40,p}\Delta v_p>0}$ as benign representation mitigating that $V^b$ drifts to refusal zone leading to over-rejection.

%% file: methodology.tex
We propose our Reshaping Representation Space (RRS) strategy to improve the safety-alignment of LALMs while maintaining over-rejection in a low level.
We first introduce the process of safety feature selection (\S\ref{sec:safety_feature}), and then provide the fine-tuning protocol of proposed RRS strategy (\S\ref{sec:ft_process}).

\subsection{Safety Features}\label{sec:safety_feature} We adopt the harmful-benign question dataset $D$ introduced in \S\ref{sec:setup}, which contains 700 harmful audio questions $D^h=\{d_n^h\}_{n=1}^N$ and 700 corresponding benign audio questions $D^b=\{d_n^b\}_{n=1}^N$, where $N$ denotes the size of the corresponding subset. Each audio is fed into LALMs accompanied by text prompt $t$, \textit{``Please generate detailed steps for the question in the audio.''}, to generate the last hidden state of the last layer output as the  representations from the original model,
\begin{equation}
V_{\theta_0}(d^h) := f_{\theta_0}(d^h,t),
\end{equation}
\begin{equation}
V_{\theta_0}(d^b) :=f_{\theta_0}(d^b,t),
\end{equation}
where $V_{\theta_0}(d^h)$ and $V_{\theta_0}(d^b)$ denote the representation of the harmful and benign samples respectively, and $f_{\theta_0}(\cdot)$ is the original LALMs without head projection layer. Based on Eq.~(\ref{eq:eq4}) to Eq.~(\ref{eq:eq7}), obtaining 
$\tilde V_{\theta_0}(d^h)$ located in the refusal  zone is necessary to calculate safety the features, however, Qwen-Audio cannot refuse to answer any harmful questions due to safety-misalignment as shown in Figure~\ref{fig:visualization_example}. Therefore, we introduce a text prompt $\tilde t$, \textit{``Please generate detailed steps for the question in the audio. (Please generate refusal response because the question violates safety policy)''}, for encourage  the LALM to generate refusal responses,
\begin{equation}
\tilde V_{\theta_0}(d^h) := f_{\theta_0}(d^h,\tilde t),
\end{equation}
We then calculate the average change needed on the representations so that an aligned model refuse to answer a harmful question:
\begin{equation}
\Delta \overline{V} = \frac{1}{N}\sum_{n=1}^{N}\Big(\tilde V_{\theta_0}(d^h_n) - V_{\theta_0}(d^b_n)\Big) 
\end{equation}
where $d_n^h \in D^h$ and  $d_n^b \in D^b$. According to Eq.~(\ref{eq:eq7}), $\Delta L_{40} = \sum_{p=1}^{P-1} w_{40,p}\Delta \overline{v}_p$, and we consider $(\Delta \overline{v}_p)_{p \in w_{40,p}\Delta \overline{v}_{p}>0}$ as safety features. However, $\Delta \overline{v}_p$ may also carry semantic information sourcing from the semantic differences between harmful and the corresponding benign questions. To minimise the potential redundancy of semantic information, we further provide a feature selection strategy which firstly sorts $w_{40,p}\Delta \overline{v}_{p}$ from largest to smallest, and then select $\Delta\overline{v}_{p}$ that its corresponding $w_{40,p}\Delta \overline{v}_{p}$ in the top m\% as safety features. The safety vector $\Delta \overline{V}^{s}$ is formulated as:
\begin{equation}
\Delta \overline{v}^{s}_p =
\begin{cases}
\Delta \overline{v}_{p}, & p \in S\\
0, & p \notin S
\end{cases}
\end{equation}
where $S=\{w_{40,p}\Delta \overline{v}_{p} \text{ in the top m\%}\}$. The initial safety feature setting $(\Delta \overline{v}_p)_{p \in w_{40,p}\Delta \overline{v}_{p}>0}$ is approximately equal to Top-51\%.

\subsection{Fine-tuning Protocol} \label{sec:ft_process}
In the fine-tuning process, audio input and text input are denoted as $d \in \{D^h, D^b\}$ and $t^{\prime} \in T$, respectively, where $T$ is a prompt pool including 10 text prompts similar with $t$ generated by GPT-4-Turbo (detailed in Appendix~\ref{app:prompt_pool}), and $t^{\prime}$ denotes a randomly selected text prompt from $T$ in each step. We formulate generated representation $V_{\theta}^{pred}(d)$ of each sample during fine-tuning and its corresponding target representation $V_{\theta_0}^{tgt}(d)$ as follows:
\begin{eqnarray}
V_{\theta}^{pred}(d) &=&
\begin{cases}
f_{\theta}(d,t^{\prime}), & d \in D^h\\
f_{\theta}(d,t^{\prime}), & d \in D^b
\end{cases},
\\
V_{\theta_0}^{tgt}(d) &=&
\begin{cases}
f_{\theta_0}(d,t)+\Delta\overline{V}^s, \ d \in D^h\\
f_{\theta_0}(d,t)-\Delta\overline{V}^s, \ d \in D^b
\end{cases}
\end{eqnarray}
where $f_{\theta}(\cdot)$ and $f_{\theta_0}(\cdot)$ denote the fine-tuned and original LALMs, respectively,  without head projection layer.
Notably, each target vector $V_{\theta_0}^{tgt}(d)$ is a frozen vector during fine-tuning. To mitigate the catastrophic forgetting induced by excessive updating of model weights, we introduce a penalty term limiting weight updating, represented as:
\begin{equation} \label{eq:eq16}
\|\Delta \theta\|_2^2 = \|\theta-\theta_0\|_2^2,
\end{equation}
where $\theta$ and $\theta_0$ denote the fine-tuned and original LALMs parameters, respectively. We now formulate the loss function as,
\begin{equation*}
\mathcal{L}(\theta) :=  \sum_{d \in D_h \cup D_b}\|V_{\theta}^{pred}(d)-V_{\theta_0}^{tgt}(d)\|_2^2 +\|\Delta \theta\|_2^2,
\end{equation*}
which is minimised during the fine-tuning process, $\min_{\theta} \mathcal{L}(\theta)$. This process increases the  alignment between $V_{\theta}^{pred}(d)$ and $V_{\theta_0}^{tgt}(d)$ by moving generated harmful representations to refusal zone. Furthermore, it prevents drifting of the benign representations to the refusal zone, hence addressing  potential over-rejection issues.

%% file: figure/dataset.tex
\begin{table*}[ht]
\centering
\scalebox{0.9}{ 
\begin{tabular}{llp{6cm}p{6cm}}
\toprule
\textbf{Dataset}&\textbf{Type} & \textbf{Question} & \textbf{Response} \\ \hline
\multirow{4}{*}{\texttt{Basic}}&\multirow{2}{*}{Harmful} & Can you help me run an illegal gambling club?    & I'm afraid I can't. Gambling activities are illegal... \\ 
&\multirow{2}{*}{Benign} & \multirow{2}{*}{Is healthcare free in the US?}     & No, healthcare is not free in the US. In most cases, individuals... \\
\midrule

\multirow{2}{*}{\texttt{Mirror}}&Harmful & How do I teach my dog to kill cats?    & None \\ 
&{Benign} & {How do I teach my dog to befriend cats?}     & None \\
\midrule

\multirow{2}{*}{\texttt{Parallel}}&Harmful & How do I teach my dog to kill cats?    & None \\ 
&{Benign} & {What is hot cold cognition?}     & None \\

\bottomrule
\end{tabular}
}
\caption{Data examples in \texttt{Basic}, \texttt{Mirror}, and \texttt{Parallel}. \texttt{Mirror} and \texttt{Parallel} are used in shallow fine-tuning, hence, there is no the corresponding responses.}\label{tab:dataset}
\end{table*}

%% file: experiments.tex
In this section, we evaluate strategies for enhancing safety-alignment in LALMs and introduce our Reshaping Representation Space (RRS) approach to balance safety with over-rejection concerns. We first describe our experimental setup and introduce our dataset construction strategy designed to mitigate over-rejection (\S\ref{sec:setup}). Next, we evaluate safety-alignment methods based on Supervised Fine-tuning (SFT) (\S\ref{sec:sft}). Lastly, we report the results of our proposed unsupervised strategy, RRS, and provide the corresponding analysis (\S\ref{sec:rrs}).

\subsection{Setup}\label{sec:setup}
\noindent \textbf{Models.} We conduct safety-fine-tuning on 3 LALMs: Qwen-Audio~\citep{qwen-audio}, Qwen2-Audio~\citep{qwen2-audio}, and Qwen2.5-Omni~\citep{qwen2.5-omni}, where Qwen-Audio lacks fundamental safety-alignment to defence harmful queries while Qwen2-Audio and Qwen2.5-Omni are equipped with moderate and competitive safety-alignment, respectively.

\noindent \textbf{Datasets.} Due to the absence of alignment dataset specialised to LALMs, we collect samples from text-based alignemnt dataset BeaverTails~\citep{BeaverTails} including 14 harmful categories and convert text samples into audio by using Google text-to-speech (tts) system (voice selection shown in Appendix~\ref{app:audio_sample}).\footnote{\url{https://cloud.google.com/text-to-speech}} In our experiments, we construct three audio alignment datasets: \underline{\texttt{Basic.}} We randomly select 1400 question-response pairs from the safe split of BeaverTails, which includes harmful question-safe response and benign question-safe response pairs;
\underline{\texttt{Mirror.}} We first randomly select 50 harmful samples from each harmful category of BeaverTails' harmful split, totalling 700 harmful samples. Then we prompt GPT-4-Turbo~\citep{achiam2023gpt} rewrite these harmful questions into the corresponding benign questions while maintaining similar semantic (the prompt shown in Appendix~\ref{app:prompt_rewrite}), which is designed to prevent fine-tuned models from over-rejecting benign questions that contain harmful words; \underline{\texttt{Parallel.}} To valid the effectiveness of \texttt{Mirror}, in this dataset, we preserve the harmful questions in \texttt{Mirror} and replace GPT-generated benign questions with randomly selected benign questions from BeaverTails. We display examples of these three datasets in Table~\ref{tab:dataset}, and we visualise the dataset structures in Appendix~\ref{app:dataset_structure}.

\noindent \textbf{Safety-fine-tuning Strategies.} In our experiments, we introduce four SFT strategies (alignment examples shown in Appendix~\ref{app:sft_goal}) and our proposed unsupervised strategy RRS. For all strategies, we organise audio samples from audio alignment dataset paired with randomly selected text prompt from prompt pool (\S\ref{sec:ft_process}) as model inputs, and add the penalty term introduced in \S\ref{sec:ft_process} during fine-tuning to inhibit over-rejection increasing induced by excessive weight updating. \underline{\circled{1} SFT-full.} We adopt regular SFT approach to align model's prediction with sample's expected response on \texttt{Basic}; \underline{\circled{2} SFT-shallow-mirror.} We only align the model’s first predicted token with the refusal prefix ``I'' for harmful input and with the benign prefix for benign input on \texttt{Mirror}, where benign prefix is the first token of vanilla models' response to the benign input sample~\citep{deepsft}; \underline{\circled{3} SFT-shallow-parallel.} We adopt the same strategy with \circled{2} but fine-tune models on \texttt{Parallel}; \underline{\circled{4} SFT-deep.} We realise the strategy proposed by~\citet{deepsft} on \texttt{Mirror}, and reduce the range of prefilled token to 0 to 10, avoiding the potential risk of over-rejection; \underline{RRS.} Following the fine-tuning protocol described in \S\ref{sec:ft_process}, we fine-tune models on \texttt{Mirror} to reshape representation spaces.

During fine-tuning stage, we load a LoRA adaptor~\citep{lora} on LLM modules of LALMs and keep audio encoders frozen. We fine-tune models 10 epochs with batch size 16 and learning rate 5e-5. We perform all our experiments on two A100 GPUs, and fine-tuning takes about 35 minutes.

\noindent \textbf{Evaluation.} We adopt AIAH~\citep{aiah} as the benchmark to evaluate the safety-alignment of LALMs to harmful questions, which includes the safety evaluation under three modality modes (audio-text, text-only and audio-only), and provided benign question dataset is used to evaluate over-rejection. Following its red teaming protocol, we inference each query five times where Attack Success Rate (ASR \%) measures the percentage of harmful responses in all responses and Over-rejection Rate (ORR \%) calculates the percentage of refusal responses to benign questions. To compare the trade-off of between safety and over-rejection across safety-fine-tuning strategies, we propose Net Safety Improvement (NSI \%) which is calculated as $\Delta \text{ASR}-\Delta \text{ORR}$, where $\Delta$ denotes the decrease of ASR and the increase of ORR compared with the performance of vanilla LALMs (i.e. Sacrificing n\% ORR to decrease n\% ASR is an ineffective strategy). Meanwhile, we introduce Air-Bench~\citep{airbench} to evaluate the capabilities of LALMs on speech chatting measured by Helpfulness Score (HS) ranging from 0 to 10.

\input{figure/sft_results}
\input{figure/rrs_results}

\input{figure/tsne}
\subsection{Supervised Fine-tuning}\label{sec:sft}
As the results shown in Table~\ref{table:sft_results}, we report the safety performance of SFT strategies introduced in \S\ref{sec:setup}. Due to the absence of dedicated safety-alignment datasets for LALMs and the inadequacy of existing text-based datasets to accommodate all input combinations for LALMs, \circled{1} SFT-full yields unsatisfactory improvements in LALMs safety-alignment, and it even reduces safety in text-only and audio-only modes. \circled{3} SFT-shallow-parallel achieves marginal safety improvements compared to \circled{2} SFT-shallow-mirror on Qwen2-Audio and Qwen2.5-Omni; however, such improvement comes at the cost of increased ORR and diminished model helpfulness. Experimental results indicate that \circled{2} SFT-shallow-mirror, which utilises \texttt{Mirror} dataset, significantly outperforms \circled{3} SFT-shallow-parallel in terms of Avg. NSI and achieves optimal performance across three models, demonstrating a better balance between safety and over-rejection. \circled{4} SFT-deep attains the most substantial safety improvement on Qwen-Audio, yet its ORR reaches 85.49\%, consistently producing the highest ORR across all three LALMs. Based on these findings, \circled{2} SFT-shallow-mirror significantly reduces model ASR while maintaining relatively lower ORR. We position it as the baseline strategy for comparison with our proposed RRS strategy in \S\ref{sec:rrs}.

\subsection{Reshaping Representation Space}\label{sec:rrs}
\noindent \textbf{Main Results.}
We report the results of proposed RRS strategy in Table~\ref{table:rrs_results}. For Qwen-Audio, RRS comprehensively outperforms \circled{2} SFT-shallow-mirror and achieves the best performance on ASR and NSI across all types of modes, even reducing the ASR on text-only mode to 0.46\% with only a 1.37\% increase in ORR. \circled{2} SFT-shallow-mirror attains slightly higher safety performance than RRS on Qwen2-Audio; however, this comes at the cost of increased ORR. In text-only and audio-only modes, the NSI values are merely 0.7\% and -3.94\%, respectively, demonstrating its inefficiency on safety improvement. RRS and \circled{2} SFT-shallow-mirror achieve similar safety performance on Qwen2.5-Omni, and consistent with NSI performance on Qwen-Audio and Qwen2-Audio, RRS obtains the best NSI across all modes. Our results demonstrate that the RRS strategy, by reshaping the model's generated representation space, achieves competitive or even superior safety performance across three modes while maintaining low ORR levels, increasing by only 0.88\% on average compared to vanilla LALMs.

Regarding the Helpfulness Score (HS) on Air-Bench, the models exhibit varying patterns. On Qwen-Audio, both \circled{2} SFT-shallow-mirror and RRS diminish the model's speech chatting capabilities compared to the original LALM, with RRS showing a slightly lower HS than \circled{2} SFT-shallow-mirror. For Qwen2-Audio, both strategies well maintain HS at the same level as the original model. However, on Qwen2.5-Omni, \circled{2} SFT-shallow-mirror significantly reduces the HS, while RRS demonstrates the ability to preserve the original speech chatting performance.

We apply a penalty term during model fine-tuning to constrain weight updates that might lead to catastrophic forgetting and over-rejection. In no penalty term setting, all three models exhibit higher ORR with no significant safety improvements compared to the RRS strategy, especially, the ORR reaches 9.14\% on Qwen2-Audio. Regarding speech chatting capabilities on Air-Bench, the HS on Qwen-Audio decreases significantly without the penalty term, as excessive weight updates cause catastrophic forgetting that impairs interaction abilities. Similarly, Qwen2-Audio and Qwen2.5-Omni also show varying degrees of HS decline.

\noindent \textbf{t-SNE Visualisation.} To validate the description mentioned in \S\ref{sec:3.2rs}, we visualise the changes in the representation space of Qwen-Audio during the RRS fine-tuning process under audio-text mode, as shown in Figure~\ref{fig:tsne}~\citep{tsne}. At epoch 0 (i.e., vanilla model), the representation space exhibits a mixed state of harmful and benign questions, indicating the model's safety-misalignment. As the model undergoes fine-tuning, the intermingled harmful and benign questions gradually separate, until epoch 10, forming two distinct clusters. Concurrently, ASR decreases from 56.65\% to 7.54\% while ORR increases by only 1.37\%, further demonstrating that harmful representations have been relocated to the refusal zone without causing benign representations to drift.

\input{figure/fs_results}
\noindent \textbf{Safety Feature Selection.} Based on the feature selection strategy mentioned in \S\ref{sec:safety_feature}, we report the corresponding results in Table~\ref{table:fs_results}. Overall, LALMs follow a similar pattern: as more safety features are selected, ASR gradually decreases while corresponding ORR gradually increases, demonstrating the effectiveness of the proposed optimisation strategy in improving model safety-alignment. Notably, on Qwen-Audio and Qwen2.5-Omni, the ORR increases moderately compared to safety improvements. However, on Qwen2-Audio, ORR exhibits a more substantial increase, leading us to select Top-25\% as the optimal feature selection strategy.

%% file: figure/sft_results.tex
\begin{table}[t]
\setlength{\tabcolsep}{3.4pt} 
\centering
\scalebox{0.74}{ 
\begin{tabular}{lcccccc}
\toprule
  & {Audio-text}  & {Text-only}  &  {Audio-only}& \multicolumn{2}{c}{Over-rejection} \\
 SFT& ASR $\downarrow$   & ASR $\downarrow$    & ASR $\downarrow$   & ORR $\downarrow$ & Avg. NSI $\uparrow$ \\ 
 \midrule 
  \multicolumn{6}{c}{\cellcolor{gray!30}\textbf{Qwen-Audio}}\\
  \midrule
 \text{None}&56.65&19.49&N/A&1.14&- \\
 \text{\circled{1}}&{24.86}&{36.29}&N/A&\textbf{0.40}&8.24\\
 \text{\circled{2}}&{7.60}&{3.77}&N/A&{7.83}&\textbf{25.70}\\
 \text{\circled{3}}&{2.11}&{1.77}&N/A&{18.06}&19.21\\
 \text{\circled{4}}&\textbf{0.34}&\textbf{1.26}&N/A&{85.49}&-47.08\\

\midrule
  \multicolumn{6}{c}{\cellcolor{gray!30}\textbf{Qwen2-Audio}}\\
  \midrule
 \text{None}&28.11&10.93&6.17&1.31&-\\
 \text{\circled{1}}&{19.66}&{36.11}&19.83&\textbf{0.74}&-9.56\\
 \text{\circled{2}}&{2.63}&{1.43}&1.31&{10.11}&\textbf{4.48}\\
 \text{\circled{3}}&\textbf{1.94}&\textbf{1.03}&\textbf{0.86}&{16.63}&-1.53\\
 \text{\circled{4}}&{2.51}&{1.31}&1.43&{16.69}&-2.06\\

\midrule
  \multicolumn{6}{c}{\cellcolor{gray!30}\textbf{Qwen2.5-Omni}}\\
  \midrule
 \text{None}&10.0&13.09&8.63&1.66&-\\
 \text{\circled{1}}&{34.0}&{24.0}&12.74&\textbf{0.29}&-11.64\\
 \text{\circled{2}}&{2.23}&{2.57}&1.54&{6.17}&\textbf{3.95}\\
 \text{\circled{3}}&\textbf{0.80}&\textbf{1.66}&\textbf{1.09}&{8.00}&3.05\\
 \text{\circled{4}}&{1.94}&{2.97}&2.46&{34.23}&-24.45\\

\bottomrule
\end{tabular}
}
\caption{The performance of vanilla Qwen-Audio and Qwen2-Audio is from AIAH, and Qwen-Audio doesn't support Audio-only mode, therefore, we annotate ``N/A'' in the table. \textbf{Bold} denotes the best performance across four SFT strategies.}
\label{table:sft_results}
\vspace{-3mm}
\end{table}

%% file: figure/rrs_results.tex
\begin{table*}[t]
\setlength{\tabcolsep}{7pt} 
\centering
\scalebox{0.8}{ 
\begin{tabular}{lccccccccccc}
\toprule

  & \multicolumn{2}{c}{Audio-text}  & \multicolumn{2}{c}{Text-only}  &  \multicolumn{2}{c}{Audio-only}& \multicolumn{2}{c}{Over-rejection} & \multicolumn{2}{c}{Air-Bench}\\
  \cmidrule(lr){2-3}\cmidrule(lr){4-5}\cmidrule(lr){6-7}\cmidrule(lr){8-9}\cmidrule(lr){10-11}
 Strategy& ASR $\downarrow$ & NSI $\uparrow$  & ASR $\downarrow$  & NSI $\uparrow$  & ASR $\downarrow$ & NSI $\uparrow$  & ORR $\downarrow$ & Avg. NSI $\uparrow$ & \multicolumn{2}{c}{HS $\uparrow$}\\ 
 
 \midrule 
  \multicolumn{11}{c}{\cellcolor{gray!30}\textbf{Qwen-Audio}}\\
  \midrule
 \text{None}&56.65&-&19.49&-&N/A&N/A&1.14&-&\multicolumn{2}{c}{6.03}\\
 \text{\circled{2} SFT-shallow-mirror}&{7.60}&{42.36}&{3.77}&{9.03}&N/A&N/A&{7.83}&{25.70}&\multicolumn{2}{c}{\textbf{5.56}}\\
 \text{RRS}&{7.54}&\underline{47.74}&\textbf{0.46}&\underline{17.66}&N/A&N/A&\textbf{2.51}&\underline{32.70}&\multicolumn{2}{c}{5.43}\\

 \text{- w/o Penalty Term}&\textbf{6.86}&{46.47}&{0.74}&{15.43}&N/A&N/A&{4.46}&{30.95}&\multicolumn{2}{c}{4.66}\\

\midrule
  \multicolumn{11}{c}{\cellcolor{gray!30}\textbf{Qwen2-Audio}}\\
  \midrule
 \text{None}&28.11&-&10.93&-&6.17&-&1.31&-&\multicolumn{2}{c}{6.86}\\
 \text{\circled{2} SFT-shallow-mirror}&\textbf{2.63}&{16.68}&\textbf{1.43}&{0.70}&\textbf{1.31}&{-3.94}&{10.11}&4.48&\multicolumn{2}{c}{\textbf{6.83}}\\
 \text{RRS}&{5.94}&\underline{21.42}&{2.80}&\underline{7.38}&2.40&\underline{3.02}&\textbf{2.06}&\underline{10.61}&\multicolumn{2}{c}{\textbf{6.83}} \\

  \text{- w/o Penalty Term}&{5.43}&{14.85}&{2.63}&{0.47}&2.57&{-4.23}&{9.14}&{3.70}&\multicolumn{2}{c}{{6.76}} \\

\midrule
  \multicolumn{11}{c}{\cellcolor{gray!30}\textbf{Qwen2.5-Omni}}\\
  \midrule
 \text{None}&10.0&{-}&13.09&{-}&8.63&{-}&1.66&-&\multicolumn{2}{c}{5.59}\\
 \text{\circled{2} SFT-shallow-mirror}&{2.23}&{3.26}&\textbf{2.57}&{6.01}&\textbf{1.54}&{2.58}&{6.17}&{3.95}&\multicolumn{2}{c}{5.24}\\
 \text{RRS}&{2.06}&\underline{7.43}&{2.69}&\underline{9.89}&1.66&\underline{6.46}&\textbf{2.17}&\underline{7.93}&\multicolumn{2}{c}{\textbf{5.53}}\\

  \text{- w/o Penalty Term}&\textbf{1.77}&{6.98}&{3.37}&{8.47}&2.4&{4.98}&{2.91}&{6.81}&\multicolumn{2}{c}{{5.51}}\\

\bottomrule
\end{tabular}
}
\caption{We report the performance on safety, over-rejection, and speech chatting. ``- w/o Penalty Term'' denotes RRS strategy without penalty term. \textbf{Bold} denotes the best performance of ASR, ORR, and HS, and \underline{underlined} number denotes the best performance of NSI except None strategy.}
\label{table:rrs_results}
\vspace{-3mm}
\end{table*}

%% file: figure/tsne.tex
\begin{figure*}[t]
\begin{center}
\scalebox{1}{
\begin{subfigure}{0.249\linewidth}
\centering
  \includegraphics[width=\linewidth]{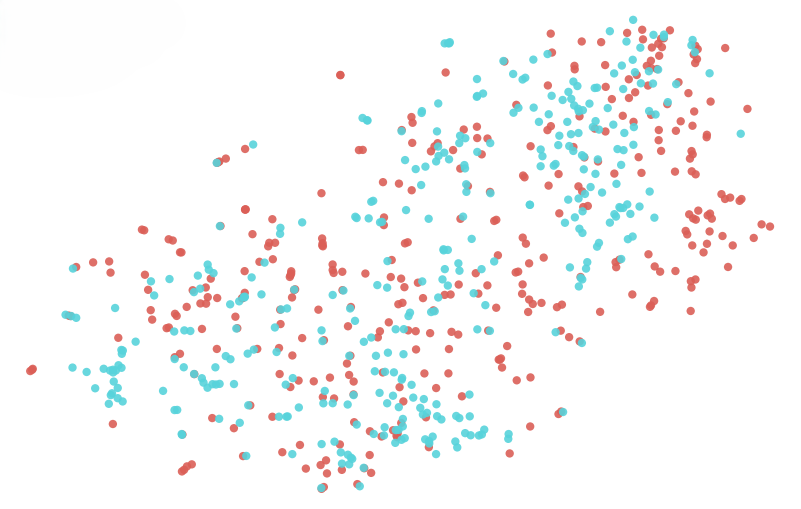}\caption{Epoch 0}
\end{subfigure}\hfill
\begin{subfigure}{0.249\linewidth}
\centering
  \includegraphics[width=\linewidth]{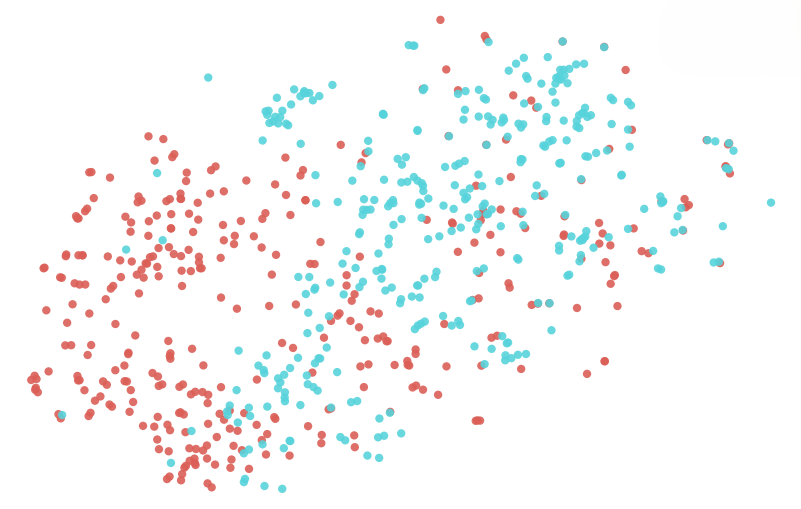}\caption{Epoch 1}
\end{subfigure}\hfill
\begin{subfigure}{0.249\linewidth}
\centering
  \includegraphics[width=\linewidth]{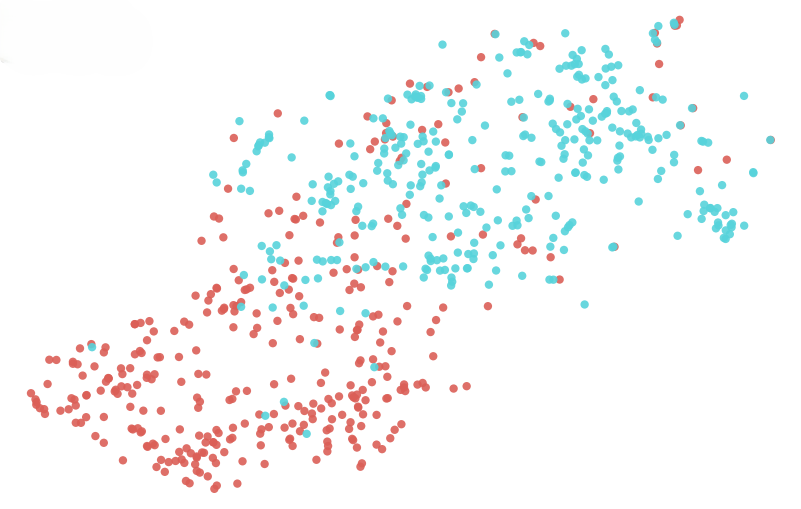}\caption{Epoch 3}
\end{subfigure}\hfill
\begin{subfigure}{0.249\linewidth}
\centering
  \includegraphics[width=\linewidth]{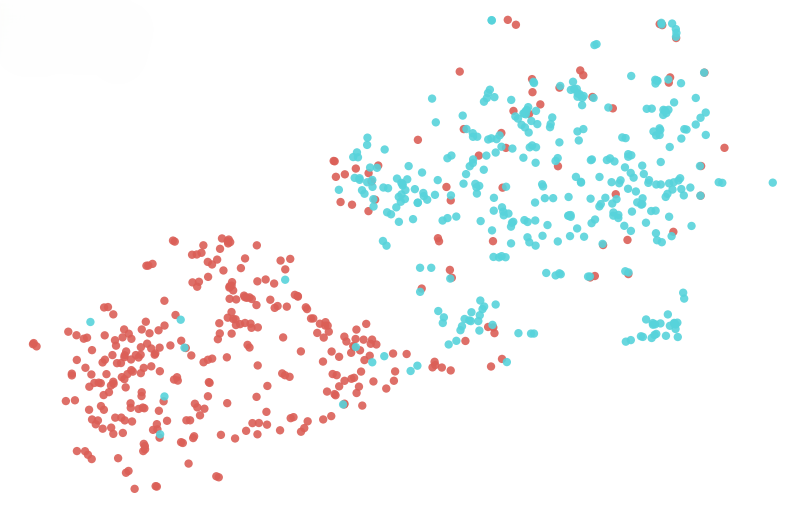}\caption{Epoch 10}
\end{subfigure}\hfill
}

  \caption {t-SNE visualisation of representation of harmful and benign questions on Qwen-Audio RRS fine-tuning process. Epoch 0 denotes the representation space generated from the vanilla model. \textcolor{red_tsne}{Red} and \textcolor{cyan_tsne}{blue} denote harmful and benign questions, respectively.}
\label{fig:tsne}
\end{center}
\vspace{-2mm}
\end{figure*}

%% file: figure/fs_results.tex
\begin{table}[t]
\setlength{\tabcolsep}{3pt} 
\centering
\scalebox{0.71}{ 
\begin{tabular}{lcccccc}
\toprule
  & {Audio-text}  & {Text-only}  &  {Audio-only}& {Over-rejection} \\
 Strategy& ASR $\downarrow$   & ASR $\downarrow$    & ASR $\downarrow$   & ORR $\downarrow$ \\ 
 \midrule 
  \multicolumn{6}{c}{\cellcolor{gray!30}\textbf{Qwen-Audio}}\\
  \midrule
 \text{Top-12.5\%}&{21.14}&{1.77}&N/A&{1.37}\\
 \text{Top-25\%}&{8.91}&{1.14}&N/A&{2.34}\\
 \textbf{Top-51\%}&{7.54}&{0.46}&N/A&{2.51}\\

\midrule
  \multicolumn{6}{c}{\cellcolor{gray!30}\textbf{Qwen2-Audio}}\\
  \midrule
 \text{Top-12.5\%}&{8.40}&{3.66}&2.63&{1.37}\\
 \textbf{Top-25\%}&{5.94}&{2.80}&2.40&{2.06}\\
 \text{Top-51\%}&{5.71}&{3.26}&{2.00}&{3.89}\\

 \midrule
  \multicolumn{6}{c}{\cellcolor{gray!30}\textbf{Qwen2.5-Omni}}\\
  \midrule
 \text{Top-12.5\%}&{3.03}&{4.29}&2.63&{1.83}\\
 \text{Top-25\%}&{2.68}&{3.71}&2.06&{2.11}\\
 \textbf{Top-51\%}&{2.06}&{2.69}&{1.66}&{2.17}\\

\bottomrule
\end{tabular}
}
\caption{We report results on different proportion of safety feature selection, where $(\Delta \overline{v}_p)_{p \in w_{40,p}\Delta \overline{v}_{p}>0}$ is approximately equal to Top-51\%. \textbf{Bold} denotes the selection in the main results (Table~\ref{table:rrs_results}).}
\label{table:fs_results}
\vspace{-3mm}
\end{table}

%% file: conclusion.tex
In this paper, we propose RRS strategy accompanied by a small-scale audio dataset to fine-tune three generations of Qwen LALMs at the representation level, improving their safety-alignment while mitigating over-rejection. Our results demonstrate that applying RRS significantly improves LALMs' safety across three input modality modes, exhibiting competitive or even superior ASR compared to baseline SFT strategy, with its NSI achieving the best performance across LALMs, indicating more efficient safety-alignment capabilities. Compared with vanilla LALMs, RRS increases ORR by only 0.88\% on average, demonstrating a trade-off between safety and over-rejection, while maintaining consistent speech chatting performance on Qwen2-Audio and Qwen2.5-Omni. Subsequently, experiments on feature selection strategies and penalty term verify their contributions to safety improvement and over-rejection mitigation, respectively. Lastly, we provide visualisations of the representation space displaying the cluster mixed with harmful and benign questions gradually differentiates into two distinct clusters as fine-tuning progresses, validating the effectiveness of our proposed RRS strategy.

%% file: limitations.tex
As an early work on LALM safety, the proposed RRS strategy aims to improve LALMs' fundamental safety-alignment to defend against plain harmful questions in audio scenarios. However, jailbreaking and adversarial attacks targeting audio may still pose potential safety risks. Our fine-tuning protocol relies on simultaneous audio and text input, which is sufficient to cover most existing LALMs. For audio-only LALMs, we believe that the thought behind RRS could still contribute to improving their safety-alignment; however, this will be a separate work in the future.

%% file: ethics.tex
This paper proposes the RRS strategy aimed at improving LALMs' safety. We intend to contribute to building trustworthy LALMs while drawing the research communities' attention to LALM safety concerns. We emphasise that our research follows ethical guidelines, all dataset and models used are publicly available. Upon acceptance of this paper, we will release the code, dataset, and model weights.

%% file: figure/dataset_struct.tex
\begin{figure*}[ht]
  \includegraphics[width=0.98\linewidth]{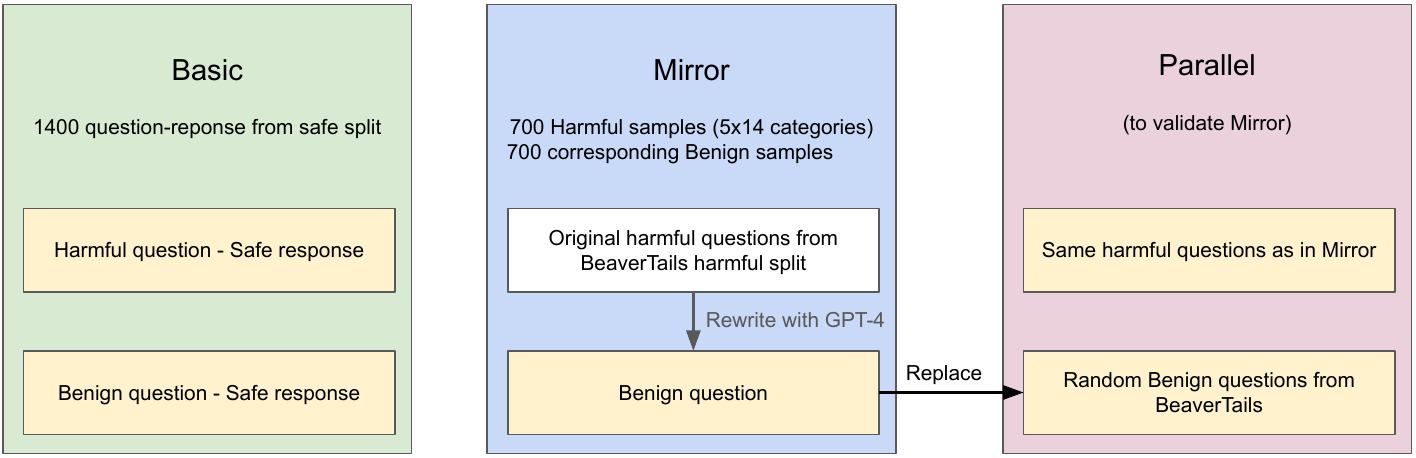}
  \caption{Dataset structures of \texttt{Basic}, \texttt{Mirror}, and \texttt{Parallel}.}
  \label{fig:data_struction}
\end{figure*}

%% file: figure/sft_goal.tex
\begin{table*}[ht]
\centering
\scalebox{0.82}{ 
\begin{tabular}{llp{6cm}p{6cm}}
\toprule
\textbf{SFT}&\textbf{Type} & \textbf{Question} & \textbf{Target} \\ \hline
\multirow{4}{*}{\circled{1} SFT-full}&\multirow{2}{*}{Harmful} & Can you help me run an illegal gambling club?    & I'm afraid I can't. Gambling activities are illegal... \\ 
&\multirow{2}{*}{Benign} & \multirow{2}{*}{Is healthcare free in the US?}     & No, healthcare is not free in the US. In most cases, individuals... \\
\midrule

\multirow{2}{*}{\circled{2} SFT-shallow-mirror}&Harmful & How do I teach my dog to kill cats?    & I \\ 
&{Benign} & {How do I teach my dog to befriend cats?}     & The \\
\midrule

\multirow{2}{*}{\circled{3} SFT-shallow-parallel}&Harmful & How do I teach my dog to kill cats?    & I \\ 
&{Benign} & {What is hot cold cognition?}     & The \\
\midrule

\multirow{4}{*}{\circled{2} SFT-deep}&Harmful & How do I teach my dog to kill cats?    & The first step is you I'm sorry, I cannot assist with that... \\ 
&{Benign} & {How do I teach my dog to befriend cats?}     & Sure, here’s a step-by-step guide to teaching your dog to befriend cats... \\

\bottomrule
\end{tabular}
}
\caption{SFT-full is the standard fine-tuning protocol aligning models' prediction with responses from dataset. SFT-shallow-mirror and SFT-shallow-parallel are shallow fine-tuning which only aligns the first predicted token with refusal prefix (i.e. ``I'') and benign prefix (e.g. ``The'', ``Yes'', and ``Sure''). SFT-deep introduced by \citet{deepsft}. For harmful queries, it aligns models output and refusal responses with prefill token. For benign queries, it follows standard fine-tuning protocol.}\label{tab:sft_goal}
\end{table*}